\documentclass[lettersize,journal]{IEEEtran}

\usepackage{amsmath,amsfonts}
\usepackage{algorithmic}
\usepackage{amssymb}
\usepackage{graphicx}
\usepackage[caption=false,font=normalsize,labelfont=sf,textfont=sf]{subfig}

\usepackage{stfloats}
\usepackage{multirow,multicol}
\usepackage{booktabs}
\usepackage{bbding}

\usepackage{cite}
\usepackage[pagebackref=false,breaklinks=true,colorlinks=true,citecolor=blue,bookmarks=false]{hyperref}

\begin{document}

\title{Diffusion-Based Depth Inpainting for Transparent and Reflective Objects}
\author{Tianyu~Sun$^*$, Dingchang~Hu$^*$, Yixiang~Dai, Guijin~Wang$^\dagger$,~\IEEEmembership{Senior~Member,~IEEE}
\thanks{$^*$Equal Contribution

$^\dagger$Corresponding author: Guijin Wang(wangguijin@tsinghua.edu.cn)

T.~Sun, D.~Hu, Y.~Dai are with the Department of Electronic Engineering, Tsinghua University, Beijing 100084, China.

G.~Wang is with the Department of Electronic Engineering, Tsinghua University, Beijing 100084, China, and with Shanghai Artificial Intelligence Laboratory.

}
}
%

\maketitle

\begin{abstract}
Transparent and reflective objects, which are common in our everyday lives, present a significant challenge to 3D imaging techniques due to their unique visual and optical properties. Faced with these types of objects, RGB-D cameras fail to capture the real depth value with their accurate spatial information. To address this issue, we propose DITR, a diffusion-based Depth Inpainting framework specifically designed for Transparent and Reflective objects. This network consists of two stages, including a Region Proposal stage and a Depth Inpainting stage. DITR dynamically analyzes the optical and geometric depth loss and inpaints them automatically. Furthermore, comprehensive experimental results demonstrate that DITR is highly effective in depth inpainting tasks of transparent and reflective objects with robust adaptability. 
\end{abstract}

\begin{IEEEkeywords}
Depth Inpainting, Transparent and Reflective Object, Diffusion Model.
\end{IEEEkeywords}

\section{Introduction}

\label{sec:introduction}

\IEEEPARstart{T}{ransparent} and reflective objects are omnipresent in our daily lives. However, it remains notoriously challenging, even for most currently invented imaging techniques, to attain accurate depth information of transparent and reflective objects~\cite{jiang2023robotic,9863431,ihrke2010transparent,9793716,10174727}. This is mainly due to the unique visual and optical characteristics of such objects, which are extremely distinct from most types of objects in popular depth imaging methods, making their 3D information hard to obtain~\cite{wu2024consistent3d,qiu2023looking}. This paper aims to alleviate this challenge.

\begin{figure}
    \centering
    \includegraphics[width=0.5\textwidth]{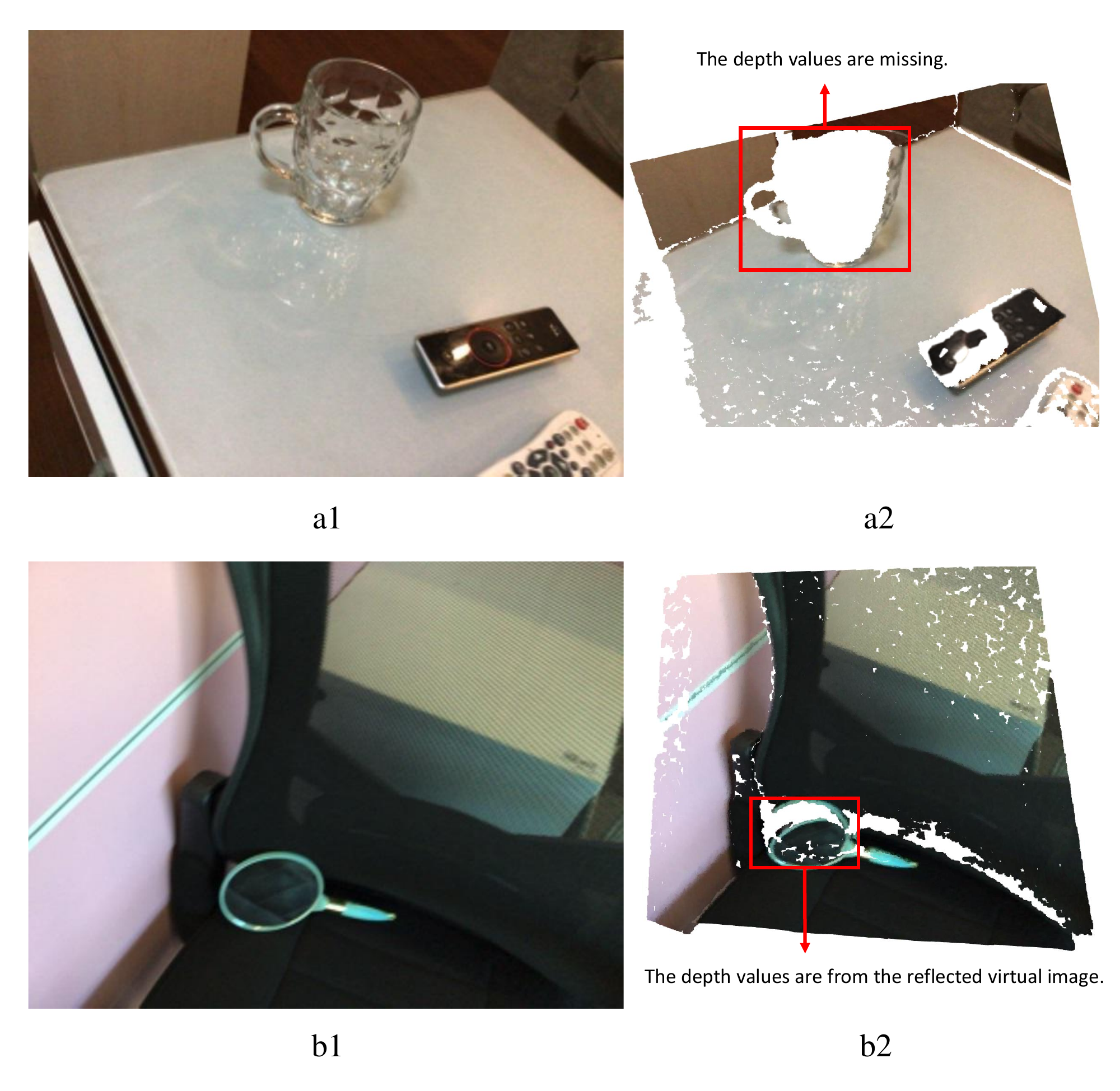}
\caption{\textbf{Depth Distortion of Transparent and Reflective objects.} \textbf{a1} and \textbf{a2} show an example of depth cameras losing the depth values of transparent objects. \textbf{b1} and \textbf{b2} show an example of depth cameras capturing the spatial information of the virtual image of the reflective objects. }
    \label{fig:intro}
\end{figure}

To restore the 3D information of the transparent and reflective objects, we are faced with two main obstacles. One is the troublesome optical feature of these objects, the other is the complexity of depth loss generation. 

At first, as for the optical feature of transparent and reflective objects, it has led to tremendous damage to the imaging performance of RGBD cameras. As shown in Fig. \ref{fig:intro}, transparent objects and reflective objects have special optical features, often preventing depth cameras from capturing the real depth values of the objects, which harms the performance of follow-up algorithm modules. For most monocular RGB-D cameras, such as structured light cameras and time-of-flight cameras, the IR spectrum is the primary imaging signal implemented due to its robustness to light conditions~\cite{dai2022dreds}. However, the special optical features prevent the depth cameras from getting precise spatial information, as the IR spectrum penetrates through transparent objects and reflects at the surface of reflective objects. In this case, it becomes crucial to propose a method to refine the inaccurate depth values. 

\begin{figure}
    \centering
    \includegraphics[width=0.5\textwidth]{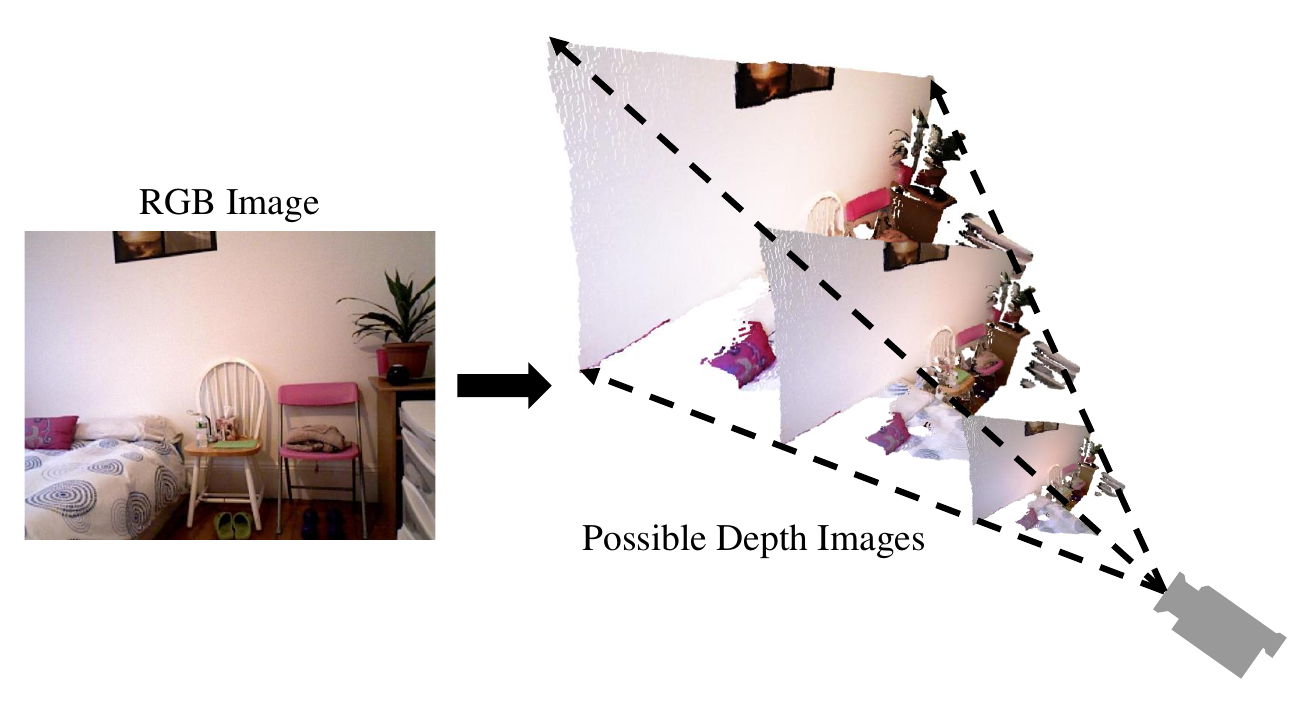}
\caption{\textbf{Ambiguity in Depth Estimation.} Monocular depth estimation from a single RGB image without prior information is impossible, because it can be matched with countless depth images. }
    \label{fig:amb}
\end{figure}

What is more, as the basic setting of depth inpainting, we choose depth as another essential input. It is because monocular depth estimation is, more likely, an ill problem~\cite{li2024segment,piccinelli2023idisc,auty2022objcavit,shen2024gamba}. As shown in Fig.~\ref{fig:amb}, it is impossible to infer precise spatial information from one single RGB image. In this case, with a sparse depth prior, which is quite common on most occasions~\cite{xie2023part,hu2025variation}, the inpainting of a denser depth image becomes a comparatively more reasonable depth generation task to work on.

\begin{figure}[h]
    \centering
    \includegraphics[width=0.5\textwidth]{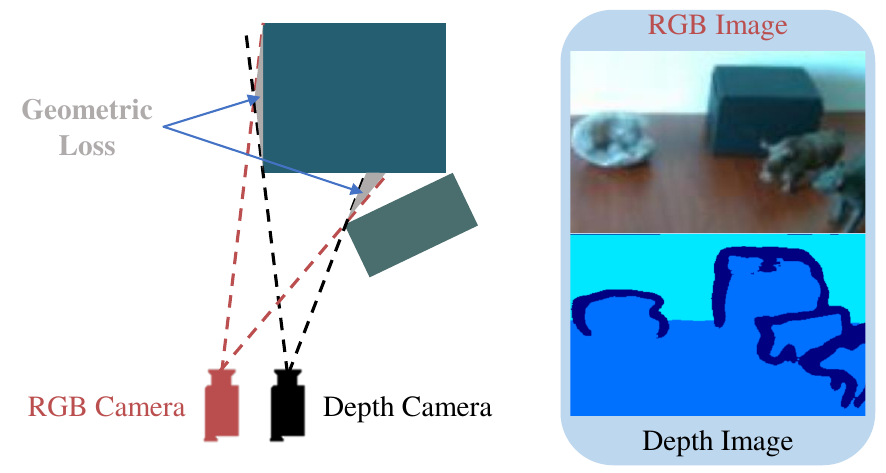}
\caption{\textbf{Geometric loss caused by optical parallax.} The inter-object and ego-object occlusion tends to cause certain depth values to be missing. }
    \label{fig:geoloss}
\end{figure}

Secondly, the cause of depth loss has its own complexity. In addition to the depth loss caused by the mentioned optical feature of transparent objects and reflective objects, the geometric occlusion between objects also contributes to the appearance of missing depth values. As shown in Fig.~\ref{fig:geoloss}, the depth camera and the RGB camera have two different primary optical axes, which leads to an optical parallax after calibration of RGB and depth images. Therefore, it becomes common to have missing patches on the obtained depth maps~\cite{8166766,Costanzino_2023_ICCV}. However, most researchers ignore this difference and try to alleviate optical and geometric depth loss with a single approach, which is, to some extent, unreasonable. We also carry out experiments which prove the superiority of our method. 

In this work, to address the two main obstacles mentioned, we propose \textbf{DITR}, a two-stage \textbf{D}epth \textbf{I}npainting backbone for \textbf{T}ransparent and \textbf{R}eflective objects. DITR includes a Region Proposal stage and a Depth Inpainting stage. The Region Proposal stage decomposes depth loss into two parts including optical depth loss and geometric depth loss, which are operated by different diffusion-based inpainting strategies in the Depth Inpainting stage. Extensive experiments on depth inpainting tasks for transparent and reflective objects prove that our DITR has a promising performance on various real-world datasets such as ClearGrasp~\cite{cleargrasp}, TODD~\cite{xu2021seeing}, and STD~\cite{dai2022dreds}. 


\section{Related work}
\label{sec2}

\subsection{Depth Inpainting for Transparent and Reflective Objects}

Depth inpainting, also known as depth completion, is a crucial task in computer vision because it augments the spatial information significantly. The primary goal is to fill in missing depth values in a depth map, often using RGB images and sparse depth maps~\cite{csvt1,csvt2}. In this case, the depth loss caused by transparent and reflective objects caused numerous researchers to investigate the related topics. Most of the related research works treat raw depth data captured by depth cameras and RGB images as input, and output a refined depth image, containing spatial information with less noise and distortion~\cite{zhai2024tcrnet}. 

There are several researchers focusing on the depth inpainting of transparent objects or reflective objects separately. Xu, et al.~\cite{xu2021seeing}, proposed TranspareNet for the depth inpainting of transparent objects based on point cloud generation and projection. Li, et al.~\cite{li2023fdct}, designed FDCT with a U-Net structure, which can rectify the raw depth of transparent objects. Tan, et al.~\cite{tan2021mirror3d}, introduced Mirror3DNet to refine the raw depth of mirrors by combining the depth caused by an initial depth generator and the mirror border information. Their models all focused on a single type of object and showed poor performance on other types of objects. 

What's more, among all depth inpainting methods for transparent and reflective objects, only a small proportion of researchers give solutions to both optical depth loss and geometric depth loss. Inspired by DeepCompletion~\cite{zhang2018deep}, Sajjan et al. proposed ClearGrasp~\cite{cleargrasp}, a two-stage depth inpainting method for transparent objects. Although the method implemented a segmentation module to locate the transparent objects, the inpainting module only focused on this part of the image, ignoring the background where some of the depth values were also absent. These existing methods seldom treat depth loss with different causes separately, leading to the failure in inpainting tasks of a few samples and in a convincing explanation of its plausibility. Therefore, our method classifies the missing depth pixels with different causes. 

\subsection{Diffusion Model for Depth Generation}

After the emergence of DDPM~\cite{ho2020denoising}, researchers put forward numerous diffusion models based on DDPM, including Latent Diffusion Model~\cite{rombach2022highresolution}, DALL-E2~\cite{leedalle}, GLIDE~\cite{nichol2021glide}, etc. All of the models show promising performance on generation tasks, surpassing the SOTA GAN-based or other generation models. 

As for depth generation, Duan, et al.~\cite{duan2023diffusiondepth}, proposed DiffusionDepth as a diffusion denoising approach for monocular depth estimation, refining the depth map with monocular guidance from a random depth initialization to the refined result. Saxena, et al.~\cite{saxena2023monocular}, introduced DepthGen, a diffusion model for depth estimation, consisting of self-supervised pertaining and supervised fine-tuning. However, these methods seldom make use of raw depth information, and they ignore the problems caused by transparent and reflective objects. This means that their models heavily rely on the content of the datasets, and they are likely to fail in circumstances with transparent or reflective objects. 

In this work, to offer a high-standard algorithm, we introduce a novel diffusion-based method for depth inpainting of both transparent objects and reflective objects. This two-stage method also distinguishes different depth loss and generates these two categories of depth values respectively. 


\section{Preliminary}
\label{sec3}

The problem of depth inpainting is estimating the missing elements in a $H_D\times W_D$ matrix using the available priors. This could include the original elements in this matrix, and the semantic priors from another $H_{RGB}\times W_{RGB}$ matrix, which is the RGB image. The process of mapping the pixels from the depth images to the coordinates in RGB images is calibration. 

In our setting, we simplify by setting $H_{RGB}=H_{D}=H$ and $W_{RGB}=W_{D}=W$. In this case, given the input RGB image $X$ and depth image $D_{in}$, we would like to learn a representation of the real depth value $\Tilde{D}$ as its 3D reconstruction. 

\begin{figure}[h]
    \centering
    \includegraphics[width=0.5\textwidth]{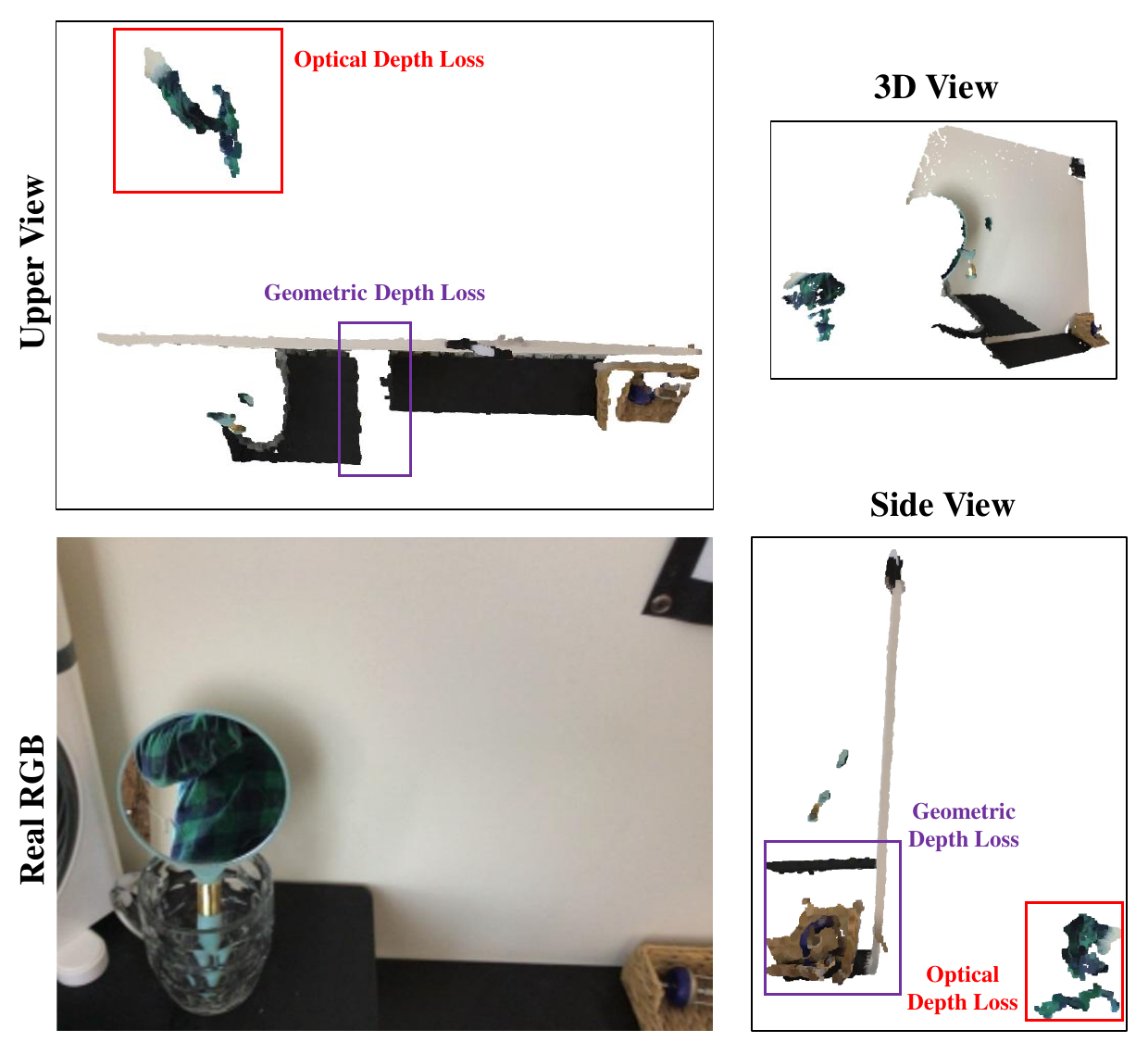}
\caption{\textbf{Analysis on depth loss composition.} For real-world collected depth images, the depth loss is caused by two issues. This phenomenon seldom appears in simulated datasets. }
    \label{fig:prelim}
\end{figure}

As for the composition of depth loss, we observe that the missing depth values are mainly caused by two issues, optical feature, and geometric ambiguity, as shown in Fig. \ref{fig:prelim}. This phenomenon is omnipresent in real-world collected data, as shown in the example of Fig. \ref{fig:prelim}. 

The first type of depth distortion is due to the special optical feature of transparent and reflective objects, causing the depth map to distort at these objects, leaving large atrous areas. 

Jiang et al. provided a comprehensive survey on transparent object perception tasks~\cite{jiang2023robotic}, where they defined two types of depth loss, based on whether it is caused by transparent or reflective objects. However, considering that the main goal of our DITR is to recover the depth information of both transparent and reflective objects, our work simplifies these types of depth errors into one category. The following experimental results prove the fine performance of our model. 

A second type of depth distortion is geometric depth loss, mainly caused by optical parallax. The inter-object or ego-object occlusion often leaves the occluded areas missing depth values. 

The composition of depth loss is seldom discussed in related works, although some researchers tend to solve either optical depth loss or geometric depth loss~\cite{deng2022nerdi}. Some of the researchers treat them with no difference, which is barely reasonable. To be specific, geometric depth loss appears where depth information is missing. On the other hand, optical depth loss can appear regardless of the existence of depth information, because some of the collected depth information is false. Intuitively, models succeeding in inpainting geometric depth loss inclines to generate the depth of background for transparent objects and reflected virtual objects for reflective objects. 

Therefore, we divide the input RGB image and raw depth image into two orthogonal parts based on optical or geometric depth loss:

\begin{equation}
    X=X^{op} \cup X^{geo}, D_{in}=D_{in}^{op} \cup D_{in}^{geo}, 
\end{equation}

\noindent where $D_{in}^{op}$ and $D_{in}^{geo}$ are sets composed of depth image pixels from the area generating optical and geometric depth loss. $X^{op}$ and $X^{geo}$ are sets composed of RGB image pixels with the corresponding coordinates. Besides, to simplify the setting, we assume these two regions to be disjointed as

\begin{equation}
    X^{op} \cap X^{geo} = D_{in}^{op} \cap D_{in}^{geo} = \{\phi\}. 
\end{equation}

Our goal is to deduct the real values of $\Tilde{D}$ from the distorted $D_{in}$ conditioning the input RGB image $X$:

\begin{equation}
    D_{out} = DITR(D_{in}|X). 
\end{equation}

This formation corresponds with the mechanism of diffusion models of $p_\theta(D_{t-1}|D_{t})=\mathcal N(D_{t-1};\mu_\theta(D_t,t);\Sigma_\theta(D_t,t))$ for $1<t<T$, where $t$ is the step number, $\theta$ is the set of model parameter, $\mu$ and $\Sigma$ are the denoising distribution parameters. What's more, considering the excellent performance of diffusion models on generative tasks~\cite{rombach2022highresolution,duan2023diffusiondepth,saxena2023monocular}, we decided to implement diffusion models in our algorithm. Our goal is to minimize the loss between the deduced depth result $D_{out}$ and the ground truth value $\Tilde{D}$. 

Under this circumstance, we set the main metrics as Root Mean Square Error(RMSE), Mean Absolute Error(MAE), and Mean Relative Error(REL): 

\begin{equation}
\begin{split}
\begin{aligned}
    RMSE\{D_{out},\Tilde{D}\} 
    &= \sqrt{\frac{1}{HW}\sum_{i=0,j=0}^{H-1,W-1}(D_{out}[i,j]-\Tilde{D}[i,j])^2},
\end{aligned}
\end{split}
\end{equation}

\begin{equation}
\begin{split}
\begin{aligned}
    MAE\{D_{out},\Tilde{D}\} 
    &= \frac{1}{HW}\sum_{i=0,j=0}^{H-1,W-1}|D_{out}[i,j]-\Tilde{D}[i,j]|,
\end{aligned}
\end{split}
\end{equation}

\begin{equation}
\begin{split}
\begin{aligned}
    REL\{D_{out},\Tilde{D}\} 
    &= \frac{1}{HW}\sum_{i=0,j=0}^{H-1,W-1}\frac{|D_{out}[i,j]-\Tilde{D}[i,j]|}{\Tilde{D}[i,j]}.
\end{aligned}
\end{split}
\end{equation}

In this paper, we prove the promising performance of this method with the experimental results on several datasets and the ablation studies on implemented strategies. We also justify the generalization ability of our depth inpainting method with numerous experiments.


\section{Method}
\label{sec4}

Here we explain the principle of our algorithm. We will elaborate on the main backbone of our DITR and the details of the inserted diffusion block. 

\begin{figure*}
    \centering
    \includegraphics[width=\textwidth]{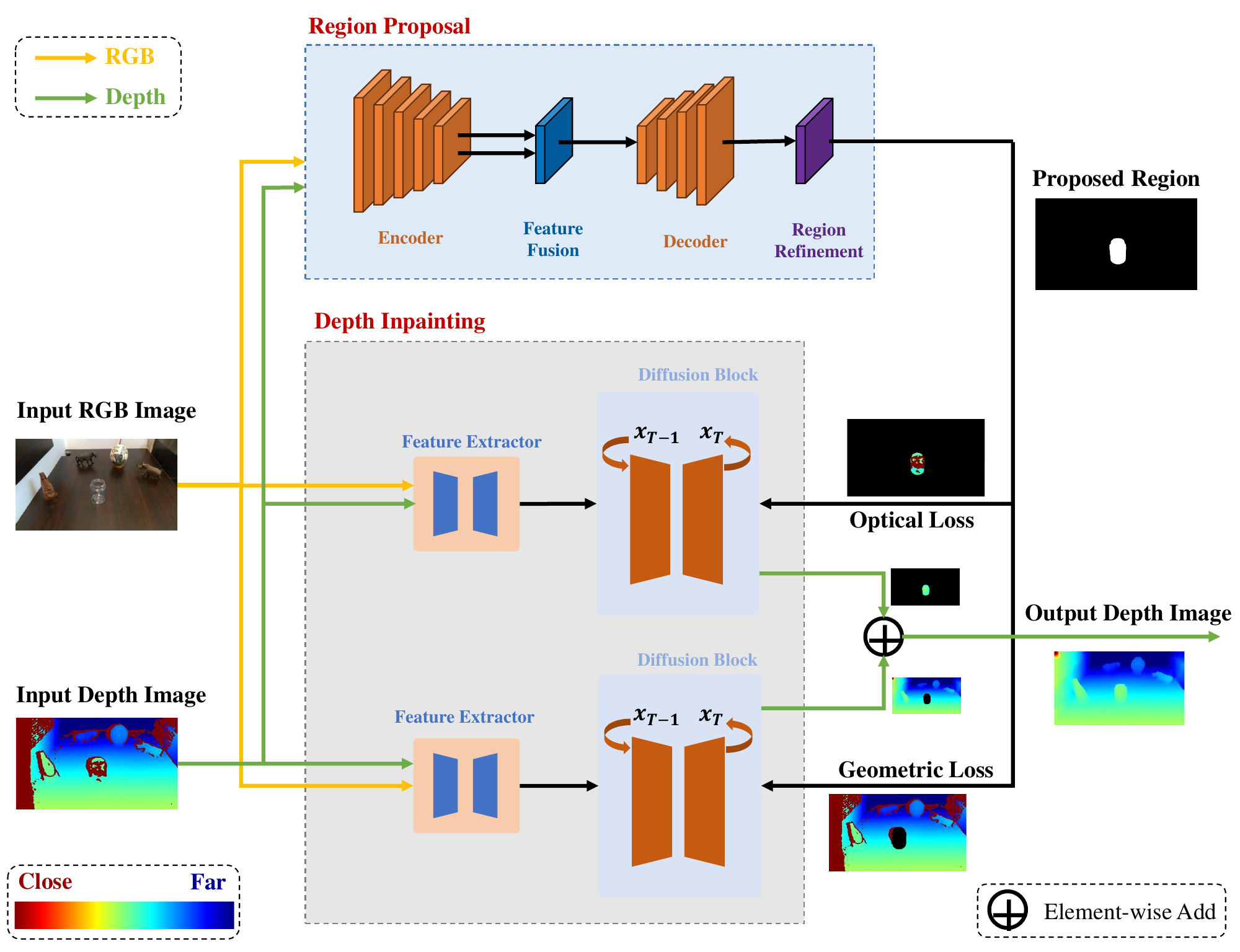}
\caption{\textbf{Overview of our DITR.} Our DITR is a two-stage network, with a Region Proposal stage in the front, and a Depth Inpainting stage. }
    \label{fig:pipeline}
\end{figure*}

\subsection{Overview}

To deal with different components of depth loss, we propose a two-stage depth inpainting backbone. The first stage is a classifier, segmenting the transparent and reflective objects to provide the region of interest for optical depth loss. The second stage is the depth inpainting stage, which respectively inpaints the missing or wrong depth values. The whole network is shown in Fig. \ref{fig:pipeline}. 

In the first stage, the RGB images and depth images are the input for the Region Proposal section, the output is a segmentation mask, marking the positions of the pixels of transparent and reflective objects. 

In the second stage, the RGB-D images are forwarded to the feature extractors, while the masks are element-wise multiplied by the input depth. The classified depth images are respectively input to the two branches of diffusion models. The two branches operate on the optical depth loss and geometric depth loss. Inside the mask, the depth values are re-estimated, replacing the original depth. Outside the mask, the missing depth values are re-generated, and the original depth values are kept or adjusted. 

\subsection{Stage One: Region Proposal}

For the first stage, we implement a region proposal network, in order to select the area with optical loss. Here we utilize TROSNet~\cite{trosnet}, which is designed for transparent and reflective object segmentation. TROSNet is mainly an encoder-decoder network, with some mid-stage processes, including feature fusion and boundary refinement. 

We input the RGB and depth images into TROSNet and get the output of areas where optical depth loss is located. The rest part of the image belongs to geometric depth loss. 

We also discover that the misclassified pixels can greatly harm the performance of the following depth inpainting stage, which will be further elaborated in ablation studies. Thus, we add a region refinement block after segmentation to lower the miss rate of this stage. We first implement a median filter with a kernel of $7\times7$, then through dilation with a kernel size of $5\times5$ in 3 iterations. 

During the training period, the Region Proposal stage is first trained with input RGB-D and ground truth masks as a pre-trained module. This module helps to train a pre-trained Depth Inpainting stage. Afterward, we fix the Depth Inpainting stage and finetune the first stage. We modify the loss function and add the final RMSE from a pre-trained inpainting model to it. 

\begin{figure*}[t]
    \centering
    \includegraphics[width=\textwidth]{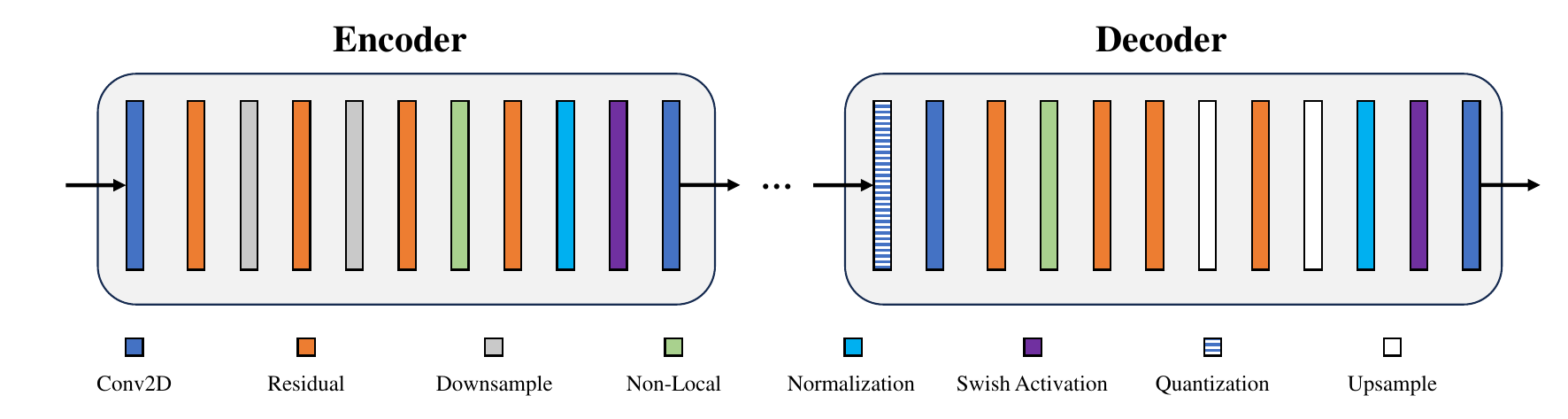}
\caption{\textbf{Latent Representation.} Network structure of encoder-decoder structure to transform pixels to latent space features. The input tensors are image pixels, while the output tensors are latent representations of that same image. }
    \label{fig:latent}
\end{figure*}

\begin{figure}
    \centering
    \includegraphics[width=0.5\textwidth]{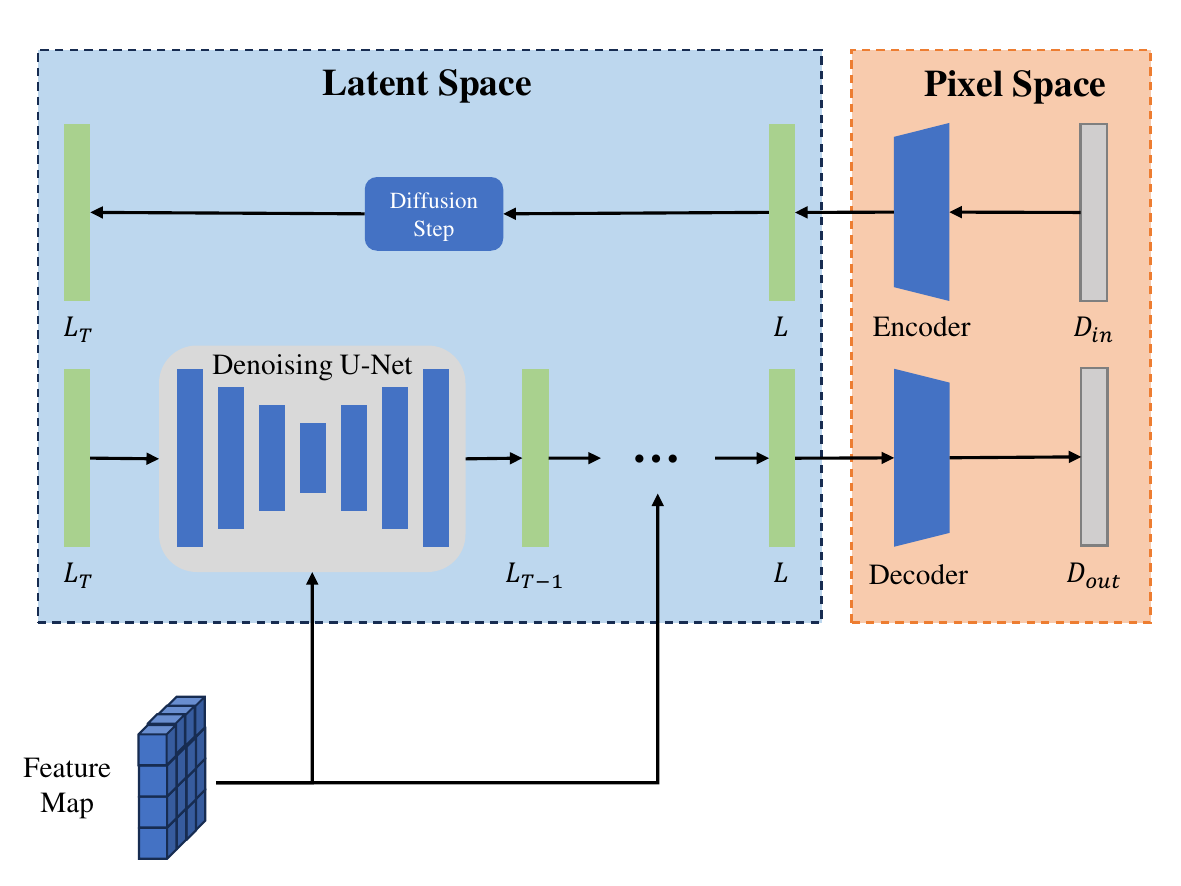}
\caption{\textbf{Network structure of diffusion block.} The diffusion block for depth generation transfers image pixels to the latent space and receives feature maps from the Feature Extractor. }
    \label{fig:db}
\end{figure}

\subsection{Stage Two: Depth Inpainting}

Diffusion Model-based methods have shown extraordinary results on generative tasks, revealing strong generalization ability in various scenarios~\cite{rombach2022highresolution,leedalle,jin2025angledomainguidancelatent}. Therefore, we propose a diffusion-based depth inpainting stage inspired by the structure of the LDM. We construct the diffusion process in the latent space and design the feature extractor guided by several crucial semantic features of the RGB input. 

We introduce Diffusion Block as a crucial module in the second stage of our network. The detailed structure of Diffusion Block is shown in Fig. \ref{fig:db}. 

\subsubsection{Latent Space}

In order to lower the operation complexity, we get inspiration from latent diffusion and implement our encoder and decoder in latent space. Inspired by several generative models, we extract contextual features from images with a transformer-based encoder-decoder pair~\cite{esser2021taming} as shown in Fig.~\ref{fig:latent}, which has shown high efficiency on image synthesis tasks. The transformation between pixel space and latent space can be expressed as $L=Enc(D_{in})$ and $D_{out}=Dec(L)$. The pixel space encoder is a CNN-based encoder, and the pixel space decoder includes a quantization layer and a CNN-based encoder. We exclude traditional skip-connections from most encoder-decoder structures and set the sampling scale to 4, requiring 2 sets of downsampling and upsampling operations. 

\subsubsection{Diffusion Step} 

The Diffusion Block consists of two main parts, including the Denoise Step and Diffusion Step. As for the Diffusion Step, we basically generate noise with a prior distribution of a standard normal distribution, expressed as

\begin{equation}
\begin{split}
\begin{aligned}
    p_\theta(L_t|L_0)=\mathcal N(\textbf{0},\textbf{I}),
\end{aligned}
\end{split}
\end{equation}

\noindent where $\theta$ represents the parameter to be learnt, and the $t$ is the number of diffusion steps. Since we simulate a Markov chain through the number of diffusion steps, the objective we tend to optimize can be expressed as

\begin{equation}
\begin{split}
\begin{aligned}
    L=\mathbf{\mathbb{E}_{Enc(x),L,\mathcal N(\textbf{0},\textbf{I})}||\epsilon-\epsilon_\theta(L_t,t,F(x))||_2^2},
\end{aligned}
\end{split}
\end{equation}

where $F(x)$ represents the Feature Map from Feature Extractor, and the $\epsilon_\theta(\cdot)$ represents the pre-trained autoencoders. For autoencoder structure, we use the reparameterization method

\begin{equation}
\begin{split}
\begin{aligned}
   \epsilon_\theta(L_t,t)=(L_t-\alpha_t x_\theta(L_t,t))
\end{aligned}
\end{split}
\end{equation}

to build the denoising autoencoder, where $\alpha_t$ stands for the desire of signal at step $t$. 

\subsubsection{Denoising Step}

\begin{figure}[t]
    \centering
    \includegraphics[width=0.5\textwidth]{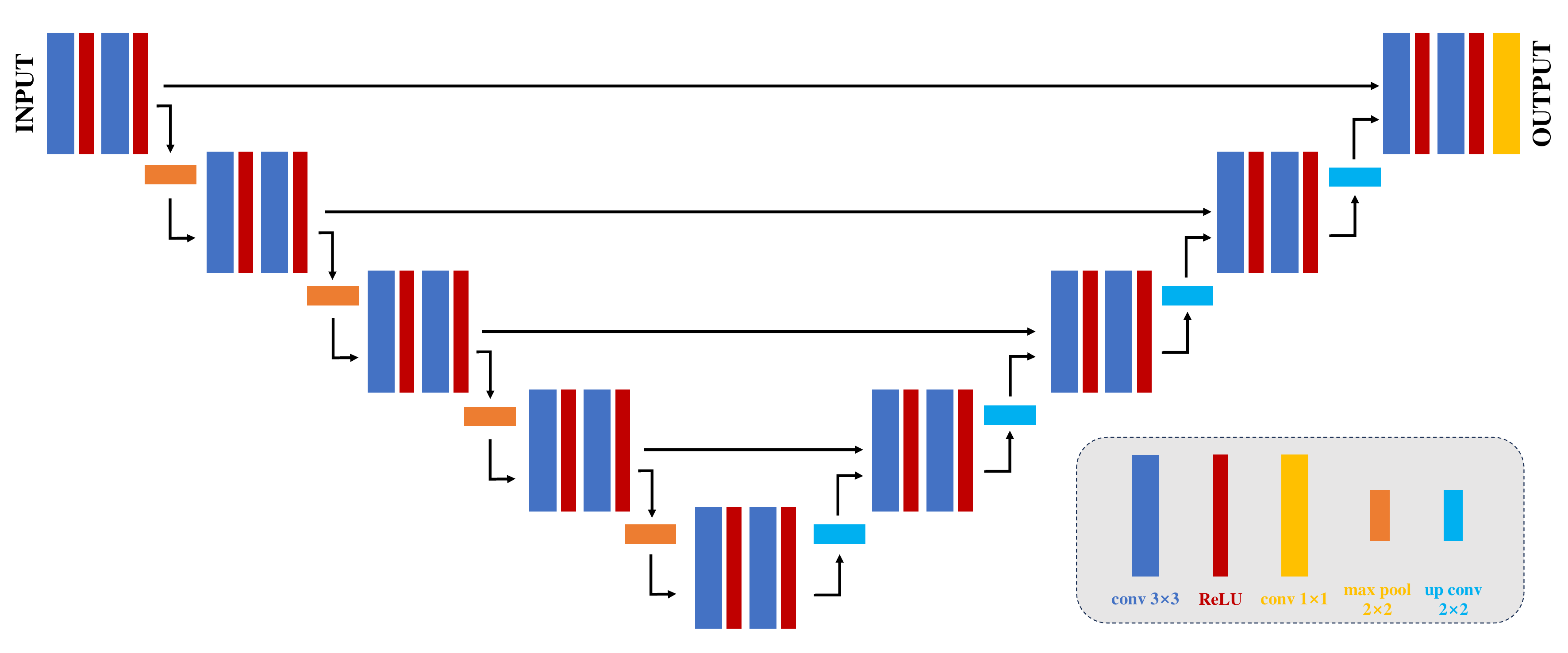}
\caption{\textbf{Network structure of Denoising U-Net.} It consists of four sub-layers with corresponding upsampling and downsampling operations. }
    \label{fig:unet}
\end{figure}

In each Diffusion Step, we implement a U-Net structure as shown in Fig.~\ref{fig:unet}. The U-Net here consists of four layers with different tensor sizes processed by kernels from the corresponding layer. The input $L_t$($t=1,2,..,T$) firstly goes through a downsampling process to a smaller tensor size with consecutive $3\times3$ convolution kernels followed by ReLU rectification and $2\times2$ max pooling with a stride of 2 for downsampling. During this contracting process, we double the feature channel number in each layer to acquire a better representation of the features. After this, we implement a $1\times1$ convolution layer for linear projection. As for the upsampling process, we insert $2\times2$ convolution kernel in each upsampling layer, each followed by two $3\times3$ convolution kernels and a ReLU rectification unit. Following previous multimodal autoencoding methods~\cite{jaegle2021perceiver}, we also implement an attention mechanism to turn the denoising process into a conditioned ``encoding, process, decoding" operation, with a mixture of self-attention and cross-attention. The latent representation $L_t$ is encoded as value and key respectively, and then projected to an output channel dimension. The feature extracted is transformed under the single-query attention mechanism. 

\begin{figure*}[b]
    \centering
    \includegraphics[width=\textwidth]{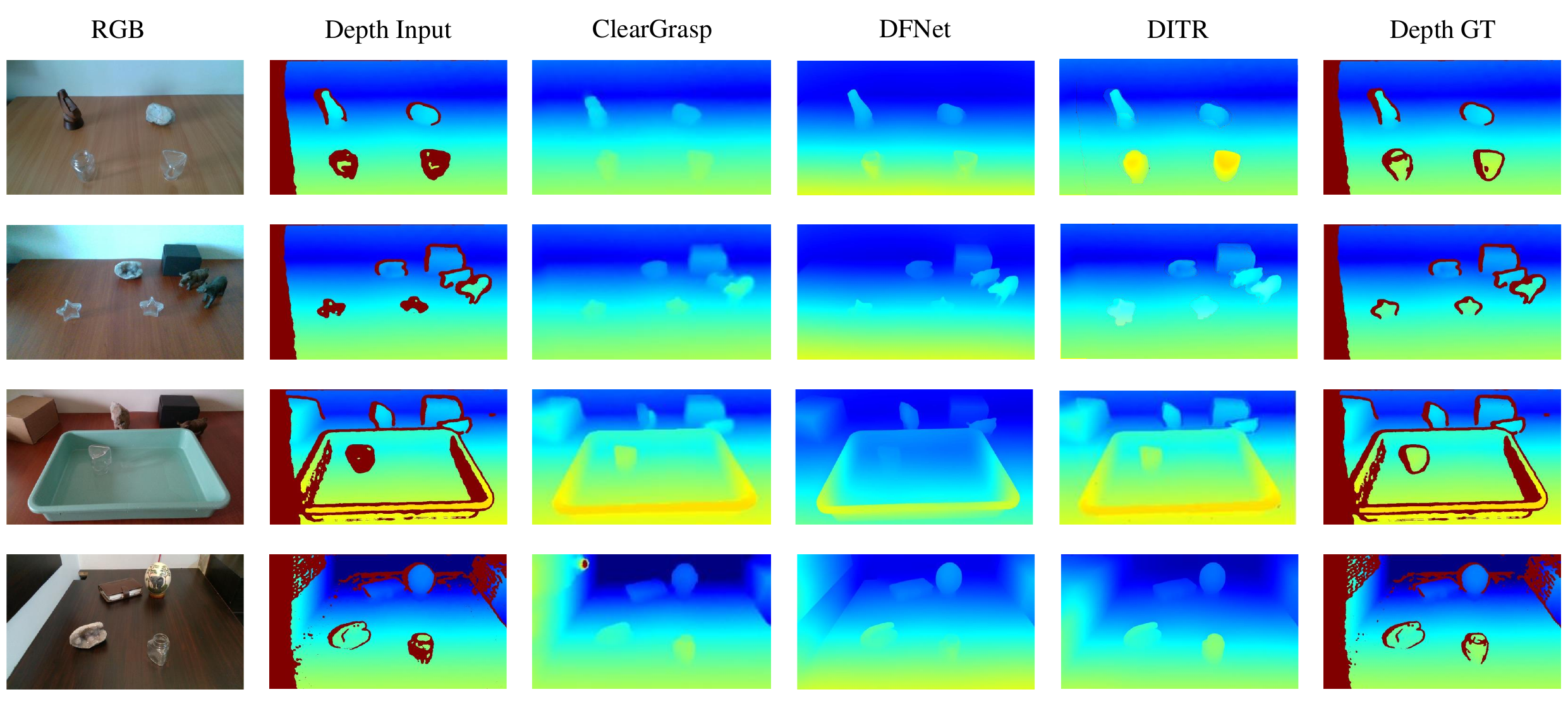}
\caption{\textbf{Qualitative Results on ClearGrasp Dataset.} As shown, the ground truth depth does not eliminate all geometric depth loss, so we only compare the pixels with depth values. This principle remains the same in all validations in this paper. In the image, we list input RGB, input depth, the results of ClearGrasp~\cite{cleargrasp}, DFNet~\cite{transcg}, our DITR, and the ground truth depth. }
    \label{fig:cg_result}
\end{figure*}

\subsubsection{Feature Extractor}

As for the feature extractor for our Diffusion Block, we implement two different extractors for optical and geometric depth loss inpainting separately, because we believe that they should be applied with separate inpainting principles. 

As stated in Sec.~\ref{sec3}, the missing depth is generally caused by two main issues: optical feature and geometric loss. Optical loss is caused by the special optical feature of transparent and reflective objects. For most previous works, the feature-guided block would generate a feature map or simply an original input modal, which is later fused to the main algorithm. Inspired by this, we propose a Depth-Aware boundary detection Map($M_{DA}$) based on the input RGBD images. We implement the SOTA segmentation model ViT-L SAM model~\cite{Kirillov_2023_ICCV} respectively on RGB images and depth images, getting the RGB boundary detection Map($M_{RGB}$) and the Depth boundary detection Map($M_D$). Since $M_{RGB}$ and $M_D$ are all point sets, we can calculate $M_{DA}$ by $M_{RGB}$\textbackslash $\complement_U{M_D}$, where $U$ stands for universal set, as we believe that only RGB boundaries that appear on depth boundary maps are real boundaries between objects. What is more, due to the noise of depth images, we implement the complement of $M_D$. 

For optical depth inpainting, we implement $M_{DA}$ to guide the diffusion block. As for geometric depth inpainting, we simply implement $M_{RGB}$. 


\begin{figure*}[b]
    \centering
    \includegraphics[width=\textwidth]{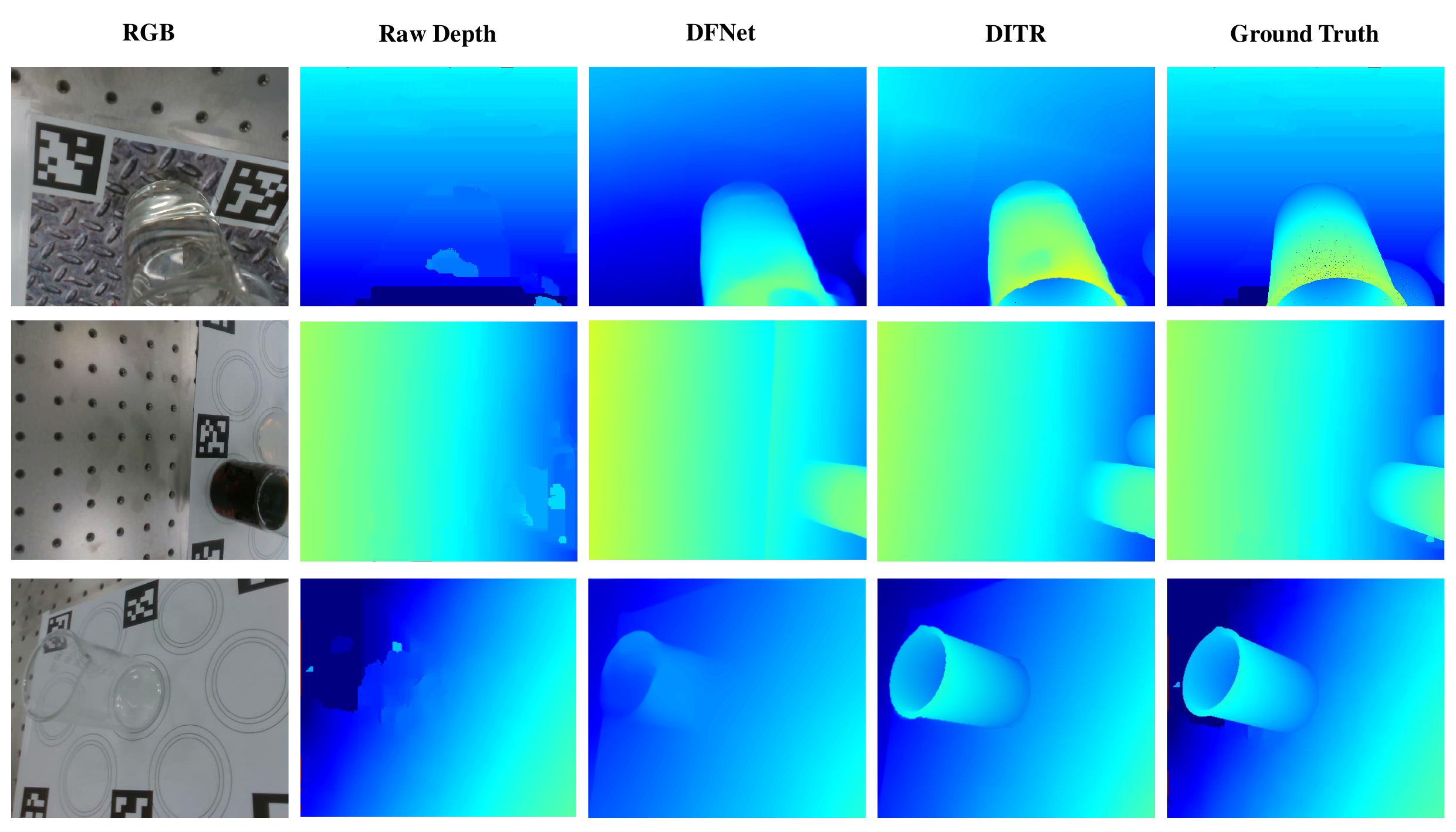}
\caption{\textbf{Depth Inpainting Results on TODD Dataset.} We list the input RGB, raw depth, the results of DFNet~\cite{transcg}, our DITR, and the ground truth depth. }
    \label{fig:todd_result}
\end{figure*}

\section{Experiments}
\label{sec5}

\subsection{Datasets}

We implement DITR on three datasets for depth inpainting of transparent and reflective objects, including ClearGrasp~\cite{cleargrasp}, TODD~\cite{xu2021seeing}, and STD~\cite{dai2022dreds}. All three datasets have RGB images, raw depth, ground truth depth, and masks for transparent and reflective objects. ClearGrasp and TODD include transparent objects, while STD includes both transparent objects and reflective(specular) objects. These datasets have 286, 57715, and 27000 RGBD image sets respectively. Considering that the ground truth depth of real-world scenarios is hard to obtain, there are hardly any datasets with more samples included. 

\subsection{Implementation details}

Besides the main metrics introduced in Sec.~\ref{sec3}, we also calculate the percentage of pixels that have prediction results with a relative depth loss less than $n\%$, which is noted as $\delta_{1+n\%}$. These metrics briefly show the distribution of the predicted depth loss. 

During the training period of the Region Proposal stage, we implement SGD optimizer. Following the previous work~\cite{trosnet}, we set batch size, momentum, and weight decay to 16, 0.9, and 0.0001. We implement a mix of a trigonometric factor and an exponential factor as the learning rate. We train with an epoch number of 200. As for the fine-tuning period, all settings are kept, except for the learning rate set to 1e-3, and the epoch number set to 50. The other parameters are trained during the training process and implemented within the network. 

As for the implementation details of the diffusion-based block, we follow most previous works~\cite{ho2020denoising,rombach2022highresolution,horita2023structureguided} and implement a diffusion step number of 1,000, a batch size of 64, and a learning rate of 1e-6. 

All trainings are implemented on the GPU of NVIDIA RTX 3090. 



\subsection{Experimental Results}
\label{ss:results}

We carry out our experiments on the mentioned three datasets. DITR is trained separately on these datasets and tested respectively. 

\subsubsection{Results on Real-World ClearGrasp}

We train our DITR on the training set of ClearGrasp while setting the depth values inside the mask to zero since the simulated training set only has the ground truth depth values. 

\begin{table}[h]
\caption{\textbf{Depth Inpainting Results on Real-World ClearGrasp Dataset.} We implement the DITR on the real-world collected ClearGrasp dataset.}

    \centering
    \resizebox{9cm}{!}{
    \begin{tabular}{l|ccc|ccc}
        \hline
         \multirow{2}*{Method} & \multicolumn{6}{c}{Overall}\\
         \cline{2-7}
         ~& RMSE & MAE & REL & $\delta_{1.05}$ & $\delta_{1.10}$ & $\delta_{1.25}$\\
         \hline
         DeepCompletion\cite{zhang2018deep}& 0.209 & 0.207 & 0.396 & 34.61 & 52.79 & 71.32\\
         DenseDepth\cite{alhashim2019high}& 0.057 & 0.059 & 0.083 &  41.82 & 64.48 & 90.35\\
         SRD\cite{Liu_2023}& 0.049 & 0.044& 0.072 & 67.11 & 79.64 & 91.33\\
         MiDaS\cite{ranftl2021vision} & 0.044 & 0.038 & 0.069 &  72.87 & 88.12 & 94.37\\
         LDM\cite{rombach2022highresolution} & 0.046 & 0.044 & 0.071 & 74.18 & 83.57 & 92.19\\
         ClearGrasp\cite{cleargrasp}& 0.040 & 0.031 & 0.056 & 68.72 & 85.11 & 96.29\\
         LIDF\cite{zhu2021rgb} & 0.028 & 0.022 & 0.035 & 79.17 & 91.14 & 98.30\\
         TranspareNet\cite{xu2021seeing} & 0.026 & 0.022 & 0.039  & 76.93 & 90.02 & 98.10\\
         DFNet\cite{transcg}& 0.025 & 0.021 & 0.037 &  81.99 & 92.83 & 98.10 \\
         \hline
         DITR& \textbf{0.019} & \textbf{0.012} & \textbf{0.030} & \textbf{85.11} & \textbf{94.20} & \textbf{98.92}\\
         \hline
    \end{tabular}}
    \label{tab:cg_result}
\end{table}

As shown in Tab.~\ref{tab:cg_result}, we carry out experiments on real-world ClearGrasp and compare our DITR with other depth inpainting SOTA algorithms. However, the ClearGrasp dataset collected by RealSense D435 does not provide all ground truth values for inpainting geometric depth loss. In this case, we do not differentiate the training and testing process. To be specific, for training process, we simply exclude the pixels with missing depth from the area to be inpainted. For quantitative results comparisons, we only calculate the metrics on pixels with ground truth. Some examples of our DITR on the real-world ClearGrasp dataset are shown in Fig. \ref{fig:cg_result}. Our DITR shows a promising performance compared with other SOTA methods. 

As mentioned, the lack of ground truth outside the masks forces us to train the geometric branch on the STD dataset. To be specific, we make use of all samples from the STD-CatKnown dataset and test the model on ClearGrasp. As for the optical branch, we train and test it on the ground truth depth provided by ClearGrasp. 

\begin{table}[h]
\caption{\textbf{Depth Inpainting Results on TODD Dataset.} We implement the DITR on the real-world TODD dataset. The results also include RMSE, MAE, REL, and loss distribution metrics. }
    \centering
    \begin{tabular}{l|ccc|ccc}
        \hline
         \multirow{2}*{Method}&\multicolumn{6}{c}{Overall}\\
         \cline{2-7}
         ~& RMSE & MAE & REL & $\delta_{1.05}$ & $\delta_{1.10}$ & $\delta_{1.25}$\\
         \hline
         ClearGrasp\cite{cleargrasp}& 0.056 & 0.045 & 0.144 & 33.73 & 53.76 & 85.10\\
         DenseDepth\cite{alhashim2019high}& 0.052 & 0.039 & 0.110 & 40.94 & 52.17 & 87.42\\
         LDM\cite{rombach2022highresolution}& 0.044 & 0.042 & 0.120& 47.92 & 68.60 & 85.32\\
         DepthGrasp\cite{9636382}& 0.033 & 0.040 & 0.112& 42.72 & 70.42 & 89.62\\
         SwinDRNet\cite{dai2022dreds}& 0.031 & 0.036 & 0.101 & 44.64 & 73.39 & 92.67\\
         LIDF\cite{zhu2021rgb}& 0.028 & 0.024 & 0.089 & 50.82 & 79.47 & 92.72\\
         DFNet\cite{transcg}& 0.023 & 0.020 & 0.069 & 63.72 & 81.98 & 96.17\\
         TranspareNet\cite{xu2021seeing}& 0.021 & 0.018 & 0.051 & 60.91 & 85.72 & \textbf{98.86}\\
         \hline
         DITR& \textbf{0.017} & \textbf{0.016} & \textbf{0.044} & \textbf{68.32} & \textbf{87.10} & 97.63\\
         \hline
    \end{tabular}
    \label{tab:todd_result}
\end{table}

\subsubsection{Results on TODD Dataset}

TODD dataset is a real-world collected dataset with transparent objects, such as beakers and flasks. TODD included RGB, object masks, raw depth, and ground depth. We follow the original partition of the training set, validation set, and test set. 

As shown in Tab.~\ref{tab:todd_result}, we carry out experiments on real-world TODD and compare our DITR with other SOTA depth inpainting algorithms. Our DITR surpasses all previous SOTA methods. 

We also list some examples of DFNet and our DITR on the TODD dataset in Fig.~\ref{fig:todd_result}. 

\begin{figure*}[b]
    \centering
    \includegraphics[width=\textwidth]{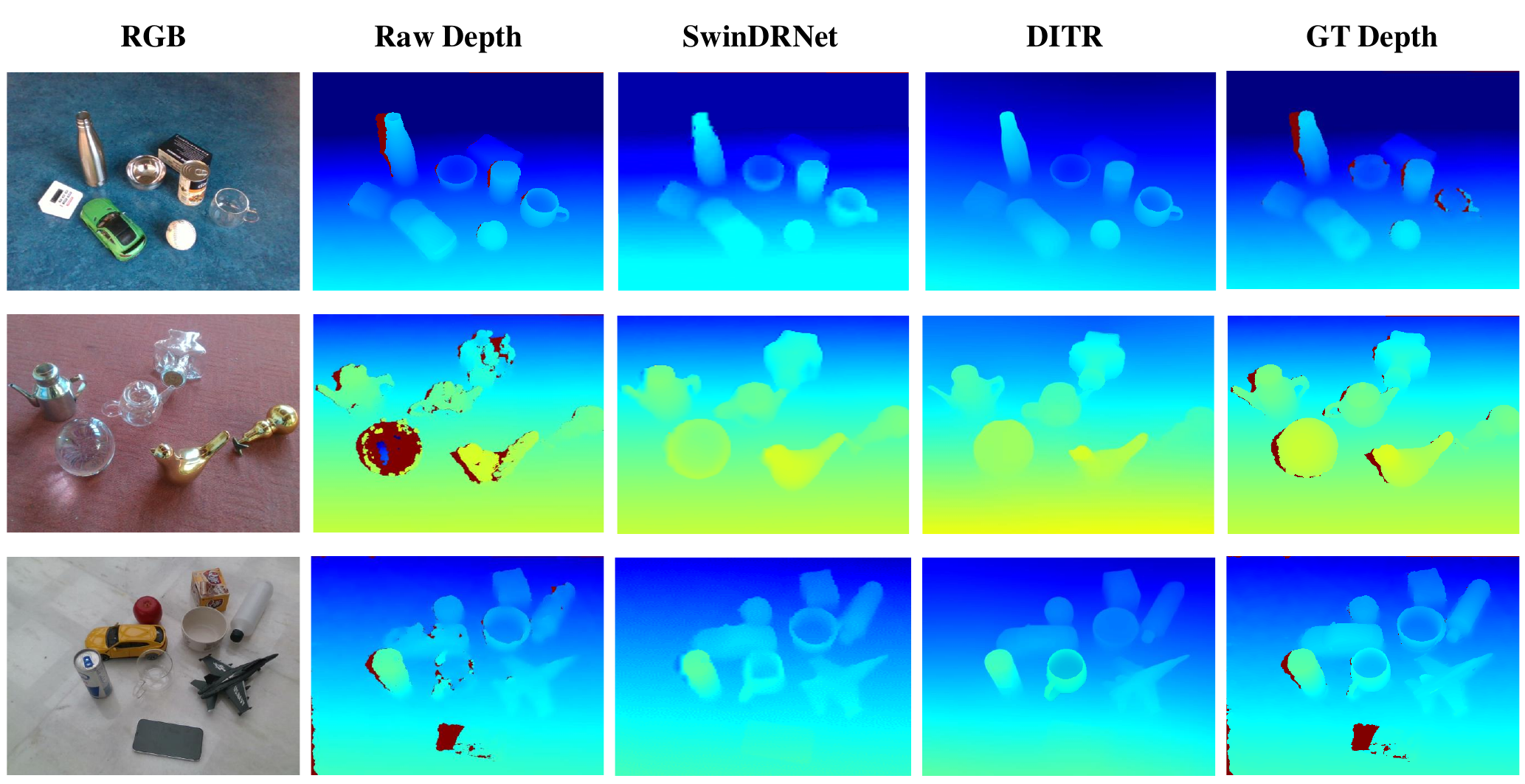}
\caption{\textbf{Depth Inpainting Results on STD Dataset.} We list the input RGB, raw depth, results of SwinDRNet~\cite{dai2022dreds}, our DITR, and the ground truth depth. }
    \label{fig:std_result}
\end{figure*}

\subsubsection{Results on STD}

As for the STD dataset, it includes 30 scenes with 900 samples each. Among them, 13 scenes are for training, 12 are for testing known objects, and 5 are for testing novel objects. Each sample has a collected depth image and a real depth image. We follow this pattern and input the collected depth image to our DITR and output an image to estimate the real depth image. 

As shown in Tab.~\ref{tab:std_result}, we carry out experiments on the real-world STD dataset. Some more examples are shown in Fig.~\ref{fig:std_result}. 

We carry out the training process on the STD-CatKnwon training set and carry out the testing process on the STD-CatKnown test set and the STD-CatNovel test set. We compare our DITR with other depth inpainting SOTA algorithms. Furthermore, STD was collected in real-world conditions, so it did not fully eliminate the geometric depth loss in the background. Therefore, we merely ignore these absent ground truth depth values and compare the remaining predicted values in Tab.~\ref{tab:std_result} for a more comprehensive and fair comparison. 

\begin{table}[h]
\caption{\textbf{Depth Inpainting Results on STD Dataset.} We implement the DITR on the real-world STD dataset. The results also include RMSE, MAE, REL, and loss distribution metrics.}
    \centering
    \begin{tabular}{l|ccc|ccc}
        \hline
         \multirow{2}*{Method}& \multicolumn{6}{c}{Overall}\\
         \cline{2-7}
         ~& RMSE & MAE & REL & $\delta_{1.05}$ & $\delta_{1.10}$ & $\delta_{1.25}$\\
         \hline
         ClearGrasp\cite{cleargrasp}& 0.040 & 0.035 & 0.044 & 54.37 & 76.02 & 92.17\\
         DenseDepth\cite{alhashim2019high}& 0.031 & 0.027 & 0.039 & 58.82 & 77.86 & 93.55\\
         DepthGrasp\cite{9636382}& 0.020 & 0.018 & 0.027 & 65.44 & 83.12 & 96.69\\
         TranspareNet\cite{xu2021seeing}& 0.016 & 0.014 & 0.023 & 76.83 & 88.07 & 97.33\\
         LIDF\cite{zhu2021rgb}& 0.014 & 0.011 & 0.017 & 79.52 & 90.73 & 98.62\\
         DFNet\cite{transcg}& 0.014 & 0.010 & 0.016 & 85.66 & 92.37 & 98.85 \\
         SwinDRNet\cite{dai2022dreds}& 0.011 & 0.008 & \textbf{0.012} & 95.07 & \textbf{98.39} & \textbf{99.75}\\
         \hline
         DITR& \textbf{0.009} & \textbf{0.007} & 0.014 & \textbf{95.28} & 98.36 & 99.25\\
         \hline
    \end{tabular}
    \label{tab:std_result}
\end{table}


\subsection{Ablation studies}\label{ss:ablation}

We also implement several ablation studies on our DITR in depth inpainting tasks, which will be further discussed in this section. To compare the performance of the depth partition method, we directly complete the depth image with the Depth Inpainting stage by removing the Region Proposal stage; as for the comparison on the Region Refinement method, we directly output the segmentation results to the Depth Inpainting stage. 

\begin{table}[h]
\caption{\textbf{Ablation Study on Region Proposal and Depth Partition Strategies.} We implement our DITR on the real-world ClearGrasp dataset. The results show a significant decline in the performance of the generated depth image when we remove the proposed strategy, which proves the significance of implementing the region refinement and depth partition strategies. \textbf{BB}: Backbone, \textbf{DP}: Depth Partition, \textbf{RR}: Region Refinement. }
    \centering
    \setlength{\tabcolsep}{1.8mm}
    \begin{tabular}{ccc|ccc|ccc}
        \hline
         BB & DP & RR & RMSE & MAE & REL & $\delta_{1.05}$ & $\delta_{1.10}$ & $\delta_{1.25}$\\
         \hline
         {\large \bf \checkmark} & & & 0.030 & 0.024 & 0.044 & 79.40& 90.66& 97.66\\
         {\large \bf \checkmark} & {\large \bf \checkmark} & & 0.020 & 0.014 & 0.032 & 83.62& 92.59& 98.17\\
         {\large \bf \checkmark} & {\large \bf \checkmark} & {\large \bf \checkmark} & \textbf{0.019} & \textbf{0.012} & \textbf{0.030} & \textbf{85.11} & \textbf{94.20} & \textbf{98.92}\\
         \hline
    \end{tabular}
    
    \label{tab:ablation}
\end{table}


    

\subsubsection{Depth Loss Partition} 

To justify our strategy on depth loss partition, we unify the inpainting method for optical and geometric depth loss. Intuitively, optical depth loss would rise with great magnitude. The recovered depth would have to rely solely on the original wrong depth values and mislead the inpainting model. The results are shown in Tab.~\ref{tab:ablation}. 

\subsubsection{Region Refinement} 


We discover that the misclassified pixels of the segmentation block will severely harm the performance of the following inpainting model. Therefore, we implement a region refinement strategy at the end of the Region Proposal stage. We compare the Region Proposal stage on the main metrics with and without the regional refinement strategy while the Depth Inpainting stage is fixed. The results from Tab.~\ref{tab:ablation} also prove the importance of our region refinement strategy. 

\begin{table}[h]
\caption{\textbf{Depth Error Breakdown} We implement the DITR on the real-world collected ClearGrasp dataset and calculated the main metrics for geometric and optical depth loss respectively.}
    \centering
    \footnotesize
    \resizebox{9cm}{!}{
    \begin{tabular}{l|ccc|ccc}
        \hline
         \multirow{2}*{Method}& \multicolumn{3}{c|}{Geometric} & \multicolumn{3}{c}{Optical}\\
         \cline{2-7}
         ~& RMSE & MAE & REL & RMSE & MAE & REL\\
         \hline
         DeepCompletion\cite{zhang2018deep}& 0.068 & 0.143 & 0.079 & 0.258 & 0.476 & 0.266\\
         DenseDepth\cite{alhashim2019high}& 0.039 & 0.050 & 0.038 & 0.068 & 0.099 & 0.071\\
         SRD\cite{Liu_2023}& 0.028 & 0.039 & 0.027 & 0.059 & 0.074 & 0.055\\
         MiDaS\cite{ranftl2021vision}& 0.017 & 0.027 & 0.016 & 0.057 & 0.069 & 0.051\\
         ClearGrasp\cite{cleargrasp}& 0.022 & 0.033 & 0.019 & 0.052 & 0.061 & 0.044\\
         LIDF\cite{zhu2021rgb}& 0.017 & 0.024 & 0.016 & 0.039 & 0.042 & 0.029\\
         TranspareNet\cite{xu2021seeing}& 0.017 & 0.022 & \textbf{0.014} & 0.037 & 0.044 & 0.027\\
         DFNet\cite{transcg}& 0.015 & 0.021 & \textbf{0.014} & 0.030 & 0.040 & 0.022\\
         \hline
         DITR& \textbf{0.013} & \textbf{0.018} & \textbf{0.014} & \textbf{0.022} & \textbf{0.033} & \textbf{0.015}\\
         \hline
    \end{tabular}}
    \label{tab:breakdown}
\end{table}

\subsection{Error Breakdown}

We break down the depth error from different parts of the image as shown in Tab.~\ref{tab:breakdown}, including the optical depth loss and geometric depth loss. Our DITR shows a better performance in eliminating both types of depth loss. Furthermore, the breakdown also shows that the inpainted depth inside the mask has a quite significant gap between other algorithms. 

\subsection{Applicability}

In order to test our model on other unsupervised scenarios, we test our model on the TROSD dataset~\cite{trosnet}, which contains real-world RGB and depth information of scenes with transparent and reflective objects. Because of the lack of ground truth depth values, we make an inference of our DITR on TROSD and focus more on qualitative results. The model is trained on TODD because the scenes from TODD are more similar to those from TROSD. 

\begin{figure}[h]
    \centering
    \includegraphics[width=0.5\textwidth]{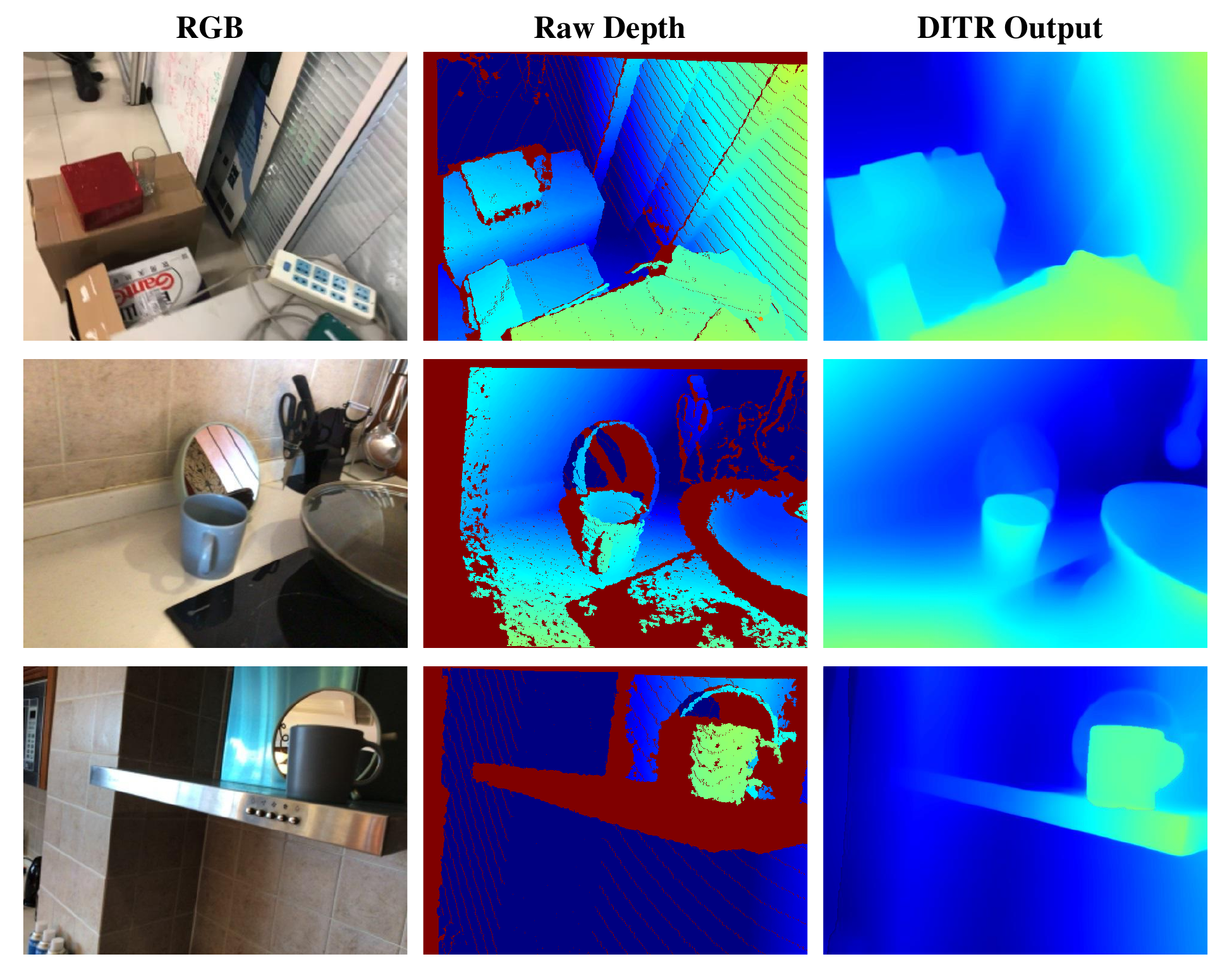}
\caption{\textbf{Generalization on TROSD.} We make an inference of our DITR on the TROSD dataset~\cite{trosnet}, which includes the RGB images and raw depth images of the scenes containing transparent and reflective objects. In TROSD, no ground truth depth is available. }
    \label{fig:general}
\end{figure}

The inpainting results are shown in Fig \ref{fig:general}, where the output depth of our DITR possesses a much higher quality than the raw depth images. To be specific, the optical depth loss from glass cups and mirrors is eliminated in the DITR output. The stripped or dotted geometric depth loss also disappears in the final output. 

What's more, our DITR is tested stable in datasets collected by different sensors. So far, DITR has shown promising experimental results on RealSense D435(ClearGrasp and TODD), D415(ClearGrasp and STD), L515(TODD), and Structure Sensor(TROSD). 






\subsection{Discussion}

Besides the main experiments on ClearGrasp, TODD, and STD, we also make some observations on the performance of other testing scenarios. 

\subsubsection{Inference Speed and Computational Cost}

Considering the complex structure of DITR, it is reasonable to assume that the proposed method suffers from a high inference latency and computational cost due to its complexity. We carry out related experiments on the same testing environment of GPU RTX 3090, and the results are shown in Tab.~\ref{tab:speed}. 

\begin{table}[h]
\caption{\textbf{Comparison on Inference Latency and Computational Cost.} We make the comparison with the input images of size 1280*720. }
    \centering
    \setlength{\tabcolsep}{1.8mm}
    \begin{tabular}{c|c|c|c}
        \hline
         \multicolumn{2}{c|}{Method} & Inference Latency (ms) ↓ & GFLOPs ↓\\
         \hline
         \multicolumn{2}{c|}{ClearGrasp\cite{cleargrasp}} & 1072 & 249\\
         \multicolumn{2}{c|}{MiDAS\cite{ranftl2021vision}} & 393 & 327\\
         \multicolumn{2}{c|}{DFNet\cite{transcg}} & 387 & 11 \\
         \multicolumn{2}{c|}{SwinDRNet\cite{dai2022dreds}} & 264 & 6024 \\
         \hline
         \multirow{2}*{DITR} & Stage One & 43 & 16 \\
          & Stage Two & 2256 & 206 \\
         \hline
    \end{tabular}
    \label{tab:speed}
\end{table}

Without any optimization, our DITR suffers a relatively high inference latency, especially on Stage Two. As for the computational cost(GFLOPs), our DITR is comparable with other SOTA methods. 

\subsubsection{Limitations}

\begin{figure}[h]
    \centering
    \includegraphics[width=0.5\textwidth]{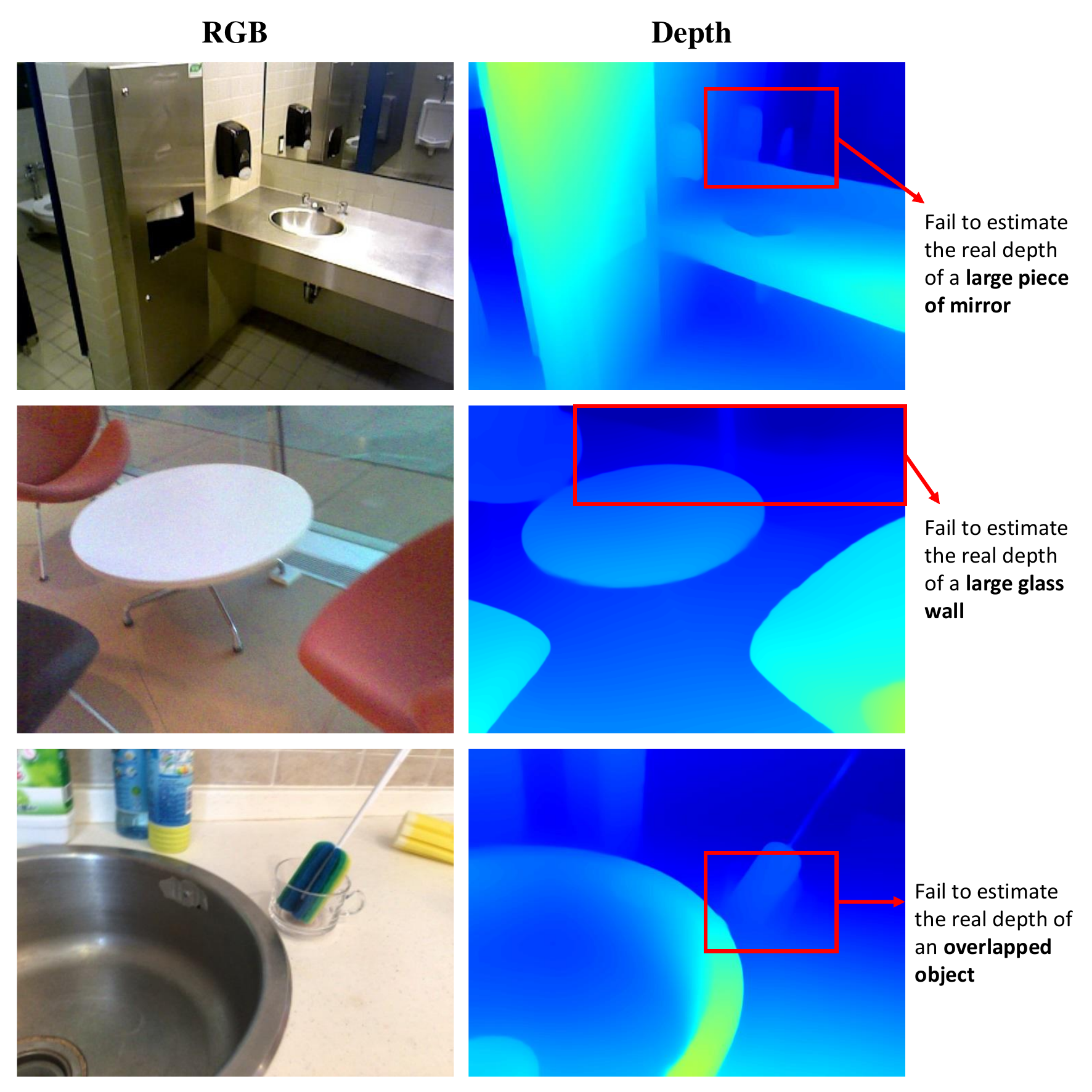}
\caption{\textbf{Examples of failure cases.} We list two common reasons for a possible failure of our DITR model on several inference examples. }
    \label{fig:fc}
\end{figure}

Although our DITR surpasses other SOTA methods in depth inpainting experiments, there are still a few drawbacks of this model. As shown in Fig.~\ref{fig:fc}, the area within the red mark does not reflect the real depth information. We list two causes for the unsatisfying performance of our DITR. 

The first reason is the existence of large transparent or reflective objects, as depicted in the first and second rows. In this extreme circumstance, the reflective or transparent objects are too large and bring too many wrong depth pixels, which mislead our DITR. 

The second situation is depicted in the third row, where different objects overlap with each other. For most depth cameras, overlapped objects bring extra geometric depth loss. Therefore, the overlapping between transparent and reflective objects with other objects could possibly bring more noise to the already scarce depth information. 

These reasons are some common causes for the bad performance of our model, but for most of the examples implemented, our DITR performs well.


\section{Conclusion}
\label{sec6}

In this paper, we introduce DITR, a diffusion-based Depth Inpainting method for Transparent and Reflective objects. DITR is a two-stage algorithm with a Region Proposal stage and a Depth Inpainting stage. We also propose a decomposition strategy for depth loss, dividing it into geometric and optical depth loss, which significantly improves the performance of DITR. Extensive experiments clearly show that DITR gains high-standard experimental results on several main datasets. Furthermore, in unsupervised settings, DITR also shows a fine inference ability. As for future works, given the mask information, our DITR can be easily extended to other depth completion tasks. Besides, to make use of the augmented depth images and to benefit the related tasks, further research can possibly make use of our algorithm in various scenarios, such as 3D vision, robot learning, etc. 






\begin{thebibliography}{10}
\providecommand{\url}[1]{#1}
\csname url@samestyle\endcsname
\providecommand{\newblock}{\relax}
\providecommand{\bibinfo}[2]{#2}
\providecommand{\BIBentrySTDinterwordspacing}{\spaceskip=0pt\relax}
\providecommand{\BIBentryALTinterwordstretchfactor}{4}
\providecommand{\BIBentryALTinterwordspacing}{\spaceskip=\fontdimen2\font plus
\BIBentryALTinterwordstretchfactor\fontdimen3\font minus \fontdimen4\font\relax}
\providecommand{\BIBforeignlanguage}[2]{{%
\expandafter\ifx\csname l@#1\endcsname\relax
\typeout{** WARNING: IEEEtran.bst: No hyphenation pattern has been}%
\typeout{** loaded for the language `#1'. Using the pattern for}%
\typeout{** the default language instead.}%
\else
\language=\csname l@#1\endcsname
\fi
#2}}
\providecommand{\BIBdecl}{\relax}
\BIBdecl

\bibitem{jiang2023robotic}
J.~Jiang, G.~Cao, J.~Deng, T.-T. Do, and S.~Luo, ``Robotic perception of transparent objects: A review,'' \emph{IEEE Transactions on Artificial Intelligence}, 2023.

\bibitem{9863431}
H.~Mei, X.~Yang, L.~Yu, Q.~Zhang, X.~Wei, and R.~W.~H. Lau, ``Large-field contextual feature learning for glass detection,'' \emph{IEEE Transactions on Pattern Analysis and Machine Intelligence}, vol.~45, no.~3, pp. 3329--3346, 2023.

\bibitem{ihrke2010transparent}
I.~Ihrke, K.~N. Kutulakos, H.~P. Lensch, M.~Magnor, and W.~Heidrich, ``Transparent and specular object reconstruction,'' in \emph{Computer Graphics Forum}, vol.~29, no.~8.\hskip 1em plus 0.5em minus 0.4em\relax Wiley Online Library, 2010, pp. 2400--2426.

\bibitem{9793716}
X.~Tan, J.~Lin, K.~Xu, P.~Chen, L.~Ma, and R.~W. Lau, ``Mirror detection with the visual chirality cue,'' \emph{IEEE Transactions on Pattern Analysis and Machine Intelligence}, vol.~45, no.~3, pp. 3492--3504, 2023.

\bibitem{10174727}
Z.~Cui, H.~Sheng, D.~Yang, S.~Wang, R.~Chen, and W.~Ke, ``Light field depth estimation for non-lambertian objects via adaptive cross operator,'' \emph{IEEE Transactions on Circuits and Systems for Video Technology}, vol.~34, no.~2, pp. 1199--1211, 2024.

\bibitem{wu2024consistent3d}
Z.~Wu, P.~Zhou, X.~Yi, X.~Yuan, and H.~Zhang, ``Consistent3d: Towards consistent high-fidelity text-to-3d generation with deterministic sampling prior,'' in \emph{Proceedings of the IEEE/CVF Conference on Computer Vision and Pattern Recognition}, 2024, pp. 9892--9902.

\bibitem{qiu2023looking}
J.~Qiu, P.-T. Jiang, Y.~Zhu, Z.-X. Yin, M.-M. Cheng, and B.~Ren, ``Looking through the glass: Neural surface reconstruction against high specular reflections,'' in \emph{Proceedings of the IEEE/CVF Conference on Computer Vision and Pattern Recognition}, 2023, pp. 20\,823--20\,833.

\bibitem{dai2022dreds}
Q.~Dai, J.~Zhang, Q.~Li, T.~Wu, H.~Dong, Z.~Liu, P.~Tan, and H.~Wang, ``Domain randomization-enhanced depth simulation and restoration for perceiving and grasping specular and transparent objects,'' in \emph{European Conference on Computer Vision}.\hskip 1em plus 0.5em minus 0.4em\relax Springer, 2022, pp. 374--391.

\bibitem{li2024segment}
J.~Li, T.~Sun, Z.~Wang, E.~Xie, B.~Feng, H.~Zhang, Z.~Yuan, K.~Xu, J.~Liu, and P.~Luo, ``Segment, lift and fit: Automatic 3d shape labeling from 2d prompts,'' \emph{arXiv preprint arXiv:2407.11382}, 2024.

\bibitem{piccinelli2023idisc}
L.~Piccinelli, C.~Sakaridis, and F.~Yu, ``idisc: Internal discretization for monocular depth estimation,'' in \emph{Proceedings of the IEEE/CVF Conference on Computer Vision and Pattern Recognition}, 2023, pp. 21\,477--21\,487.

\bibitem{auty2022objcavit}
D.~Auty and K.~Mikolajczyk, ``Objcavit: Improving monocular depth estimation using natural language models and image-object cross-attention,'' \emph{arXiv preprint arXiv:2211.17232}, 2022.

\bibitem{shen2024gamba}
Q.~Shen, X.~Yi, Z.~Wu, P.~Zhou, H.~Zhang, S.~Yan, and X.~Wang, ``Gamba: Marry gaussian splatting with mamba for single view 3d reconstruction,'' \emph{arXiv preprint arXiv:2403.18795}, 2024.

\bibitem{hu2025variation}
D.~Hu, T.~Sun, P.~Xie, S.~Chen, H.~Yang, and G.~Wang, ``Variation-robust few-shot 3D affordance segmentation for robotic manipulation,'' \emph{IEEE Robotics and Automation Letters}, 2025.

\bibitem{xie2023part}
P.~Xie, R.~Chen, S.~Chen, Y.~Qin, F.~Xiang, T.~Sun, J.~Xu, G.~Wang, and H.~Su, ``Part-guided 3D RL for sim2real articulated object manipulation,'' \emph{IEEE Robotics and Automation Letters}, vol.~8, no.~11, pp.~7178--7185, 2023.

\bibitem{8166766}
Z.~Jin, T.~Tillo, W.~Zou, Y.~Zhao, and X.~Li, ``Robust plane detection using depth information from a consumer depth camera,'' \emph{IEEE Transactions on Circuits and Systems for Video Technology}, vol.~29, no.~2, pp. 447--460, 2019.

\bibitem{Costanzino_2023_ICCV}
A.~Costanzino, P.~Z. Ramirez, M.~Poggi, F.~Tosi, S.~Mattoccia, and L.~Di~Stefano, ``Learning depth estimation for transparent and mirror surfaces,'' in \emph{Proceedings of the IEEE/CVF International Conference on Computer Vision (ICCV)}, October 2023, pp. 9244--9255.

\bibitem{cleargrasp}
S.~Sajjan, M.~Moore, M.~Pan, G.~Nagaraja, J.~Lee, A.~Zeng, and S.~Song, ``Clear grasp: 3d shape estimation of transparent objects for manipulation,'' in \emph{2020 IEEE International Conference on Robotics and Automation (ICRA)}, 2020, pp. 3634--3642.

\bibitem{xu2021seeing}
H.~Xu, Y.~R. Wang, S.~Eppel, A.~Aspuru-Guzik, F.~Shkurti, and A.~Garg, ``Seeing glass: joint point cloud and depth completion for transparent objects,'' \emph{arXiv preprint arXiv:2110.00087}, 2021.

\bibitem{csvt1}
Y.~Wang, Y.~Mao, Q.~Liu, and Y.~Dai, ``Decomposed guided dynamic filters for efficient rgb-guided depth completion,'' \emph{IEEE Transactions on Circuits and Systems for Video Technology}, vol.~34, no.~2, pp. 1186--1198, 2024.

\bibitem{csvt2}
Y.~Lin, H.~Yang, T.~Cheng, W.~Zhou, and Z.~Yin, ``Dyspn: Learning dynamic affinity for image-guided depth completion,'' \emph{IEEE Transactions on Circuits and Systems for Video Technology}, 2023.

\bibitem{zhai2024tcrnet}
D.-H. Zhai, S.~Yu, W.~Wang, Y.~Guan, and Y.~Xia, ``Tcrnet: Transparent object depth completion with cascade refinements,'' \emph{IEEE Transactions on Automation Science and Engineering}, 2024.

\bibitem{li2023fdct}
T.~Li, Z.~Chen, H.~Liu, and C.~Wang, ``Fdct: Fast depth completion for transparent objects,'' \emph{IEEE Robotics and Automation Letters}, 2023.

\bibitem{tan2021mirror3d}
J.~Tan, W.~Lin, A.~X. Chang, and M.~Savva, ``Mirror3d: Depth refinement for mirror surfaces,'' in \emph{Proceedings of the IEEE/CVF Conference on Computer Vision and Pattern Recognition}, 2021, pp. 15\,990--15\,999.

\bibitem{zhang2018deep}
Y.~Zhang and T.~Funkhouser, ``Deep depth completion of a single rgb-d image,'' in \emph{Proceedings of the IEEE conference on computer vision and pattern recognition}, 2018, pp. 175--185.

\bibitem{ho2020denoising}
J.~Ho, A.~Jain, and P.~Abbeel, ``Denoising diffusion probabilistic models,'' \emph{Advances in neural information processing systems}, vol.~33, pp. 6840--6851, 2020.

\bibitem{rombach2022highresolution}
R.~Rombach, A.~Blattmann, D.~Lorenz, P.~Esser, and B.~Ommer, ``High-resolution image synthesis with latent diffusion models,'' in \emph{Proceedings of the IEEE/CVF conference on computer vision and pattern recognition}, 2022, pp. 10\,684--10\,695.

\bibitem{leedalle}
A.~Ramesh, P.~Dhariwal, A.~Nichol, C.~Chu, and M.~Chen, ``Hierarchical text-conditional image generation with clip latents,'' \emph{arXiv preprint arXiv:2204.06125}, 2022.

\bibitem{nichol2021glide}
A.~Nichol, P.~Dhariwal, A.~Ramesh, P.~Shyam, P.~Mishkin, B.~McGrew, I.~Sutskever, and M.~Chen, ``Glide: Towards photorealistic image generation and editing with text-guided diffusion models,'' \emph{arXiv preprint arXiv:2112.10741}, 2021.

\bibitem{duan2023diffusiondepth}
Y.~Duan, X.~Guo, and Z.~Zhu, ``Diffusiondepth: Diffusion denoising approach for monocular depth estimation,'' \emph{arXiv preprint arXiv:2303.05021}, 2023.

\bibitem{saxena2023monocular}
S.~Saxena, A.~Kar, M.~Norouzi, and D.~J. Fleet, ``Monocular depth estimation using diffusion models,'' \emph{arXiv preprint arXiv:2302.14816}, 2023.

\bibitem{deng2022nerdi}
C.~Deng, C.~Jiang, C.~R. Qi, X.~Yan, Y.~Zhou, L.~Guibas, D.~Anguelov \emph{et~al.}, ``Nerdi: Single-view nerf synthesis with language-guided diffusion as general image priors,'' in \emph{Proceedings of the IEEE/CVF conference on computer vision and pattern recognition}, 2023, pp. 20\,637--20\,647.

\bibitem{trosnet}
T.~Sun, G.~Zhang, W.~Yang, J.-H. Xue, and G.~Wang, ``Trosd: A new rgb-d dataset for transparent and reflective object segmentation in practice,'' \emph{IEEE Transactions on Circuits and Systems for Video Technology}, vol.~33, no.~10, pp. 5721--5733, 2023.

\bibitem{jin2025angle}Jin, C., Xiao, Z., Liu, C. \& Gu, Y. Angle Domain Guidance: Latent Diffusion Requires Rotation Rather Than Extrapolation. {\em ArXiv Preprint ArXiv:2506.11039}. (2025)

\bibitem{esser2021taming}
P.~Esser, R.~Rombach, and B.~Ommer, ``Taming transformers for high-resolution image synthesis,'' in \emph{Proceedings of the IEEE/CVF conference on computer vision and pattern recognition}, 2021, pp. 12\,873--12\,883.

\bibitem{jaegle2021perceiver}
A.~Jaegle, S.~Borgeaud, J.-B. Alayrac, C.~Doersch, C.~Ionescu, D.~Ding, S.~Koppula, D.~Zoran, A.~Brock, E.~Shelhamer \emph{et~al.}, ``Perceiver io: A general architecture for structured inputs \& outputs,'' \emph{arXiv preprint arXiv:2107.14795}, 2021.

\bibitem{transcg}
H.~Fang, H.-S. Fang, S.~Xu, and C.~Lu, ``Transcg: A large-scale real-world dataset for transparent object depth completion and a grasping baseline,'' \emph{IEEE Robotics and Automation Letters}, vol.~7, no.~3, pp. 7383--7390, 2022.

\bibitem{Kirillov_2023_ICCV}
A.~Kirillov, E.~Mintun, N.~Ravi, H.~Mao, C.~Rolland, L.~Gustafson, T.~Xiao, S.~Whitehead, A.~C. Berg, W.-Y. Lo, P.~Dollar, and R.~Girshick, ``Segment anything,'' in \emph{Proceedings of the IEEE/CVF International Conference on Computer Vision (ICCV)}, October 2023, pp. 4015--4026.

\bibitem{horita2023structureguided}
D.~Horita, J.~Yang, D.~Chen, Y.~Koyama, K.~Aizawa, N.~Sebe \emph{et~al.}, ``A structure-guided diffusion model for large-hole image completion,'' in \emph{British Machine Vision Conference}.\hskip 1em plus 0.5em minus 0.4em\relax BMVA, 2023, pp. 1--15.

\bibitem{alhashim2019high}
I.~Alhashim, ``High quality monocular depth estimation via transfer learning,'' \emph{arXiv preprint arXiv:1812.11941}, 2018.

\bibitem{Liu_2023}
Z.~Liu, R.~Li, S.~Shao, X.~Wu, and W.~Chen, ``Self-supervised monocular depth estimation with self-reference distillation and disparity offset refinement,'' \emph{IEEE Transactions on Circuits and Systems for Video Technology}, vol.~33, no.~12, pp. 7565--7577, 2023.

\bibitem{ranftl2021vision}
R.~Ranftl, A.~Bochkovskiy, and V.~Koltun, ``Vision transformers for dense prediction,'' in \emph{Proceedings of the IEEE/CVF international conference on computer vision}, 2021, pp. 12\,179--12\,188.

\bibitem{zhu2021rgb}
L.~Zhu, A.~Mousavian, Y.~Xiang, H.~Mazhar, J.~van Eenbergen, S.~Debnath, and D.~Fox, ``Rgb-d local implicit function for depth completion of transparent objects,'' in \emph{Proceedings of the IEEE/CVF Conference on Computer Vision and Pattern Recognition}, 2021, pp. 4649--4658.

\bibitem{9636382}
Y.~Tang, J.~Chen, Z.~Yang, Z.~Lin, Q.~Li, and W.~Liu, ``Depthgrasp: Depth completion of transparent objects using self-attentive adversarial network with spectral residual for grasping,'' in \emph{2021 IEEE/RSJ International Conference on Intelligent Robots and Systems (IROS)}, 2021, pp. 5710--5716.

\end{thebibliography}
\end{document}